%% file: main.tex
\documentclass[11pt]{article}
\usepackage{acl2015}
\usepackage{times}
\usepackage{latexsym}
\usepackage{url}
\usepackage{subfigure}
\usepackage{amsmath}
\usepackage{amsthm}
\usepackage{amsfonts}
\usepackage{xspace}
\usepackage{multirow}
\usepackage{relsize}
\usepackage{covington}
\usepackage{mdwlist}
\usepackage{graphicx}
\usepackage{array}
\usepackage{colortbl}
\usepackage{color}
\usepackage{tikz}
\usetikzlibrary{patterns}
\usepackage{adjustbox}
\usepackage{hhline}
\usepackage{comment}
\newcolumntype{L}[1]{>{\raggedright\let\newline\\\arraybackslash\hspace{0pt}}m{#1}}
\newcolumntype{C}[1]{>{\centering\let\newline\\\arraybackslash\hspace{0pt}}m{#1}}
\newcolumntype{R}[1]{>{\raggedleft\let\newline\\\arraybackslash\hspace{0pt}}m{#1}}

\definecolor{ref}{rgb}{1.0,1.0,1.0} 
\definecolor{t1}{rgb}{0.25,0,0}
\definecolor{t2}{rgb}{0,0.25,0}
\definecolor{t3}{rgb}{0,0,0.25}

\newcommand{\dataset}[1]{\textsc{#1}\xspace}

\newcommand{\agentsentences}{\dataset{Sentences}}

\begin{document}

\title{Assisting Composition of Email Responses:\\ a Topic Prediction Approach}





\author{Spandana Gella\thanks{email:S.Gella@ed.ac.uk} \hspace*{1mm}, Marc Dymetman\thanks{email:Marc.Dymetman@xrce.xerox.com} \hspace*{1mm}, Jean Michel Renders\thanks{email:Jean-Michel.Renders@xrce.xerox.com} \hspace*{1mm}, Sriram Venkatapathy\thanks{email:vesriram@amazon.com} 
	\\ Xerox Research Centre Europe, France\\ }

\date{}
\maketitle
\input abstract

\input introduction
\input related

\input dataset

\input silver

\input perplexity

\input prediction

\input evaluation
\input conclusion


\clearpage

\bibliographystyle{acl}
\bibliography{references}

\end{document}

%% file: abstract.tex
\begin{abstract}

We propose an approach for helping agents compose email replies to customer requests. 
To enable that, we use LDA to extract latent topics from a collection of email exchanges. We then use these latent topics to label our data, obtaining a so-called ``silver standard'' topic labelling. We exploit this labelled set to train a classifier to: (i) predict the  topic distribution of the entire agent's email response,  based on features of the customer's email; and (ii) predict the topic distribution of the next sentence in the agent's reply, based on the customer's email features and on features of the agent's current sentence.

The experimental results on a large email collection from a contact center in the telecom domain show that the proposed approach is effective in predicting the best topic of the agent's next sentence. In {80\%} of the cases, the correct topic is present among the top five recommended topics (out of fifty possible ones). This shows the potential of this method to be applied in an interactive setting, where the agent is presented a small list of likely topics to choose from for the next sentence.
\end{abstract}

%% file: introduction.tex
\section{Introduction}

The focus of the work presented in this paper is to develop models that can help a person reply to an email query. This is very relevant in the customer care situation where agents frequently have to reply to similar queries from different customers.
Of course, those queries that are similar solicit replies that are similar as well, sharing similar {\em topic structures and vocabulary}. Hence, providing suggestions to the agent with respect to the topic structure as well as to the content, in an interactive manner, can help in the effective composition of the email reply.

Typically, in customer care centres, in order to help the agents send replies to similar queries, the agents have access to a repository of canned responses. 
In response to a query from the customer, the agent searches among appropriate canned responses, makes appropriate modifications to the text, fills in information and then sends the reply. This process is both inflexible, as well as time consuming, specially in cases where the customer query is slightly different from one of the expected queries. 

In what we are proposing here, the goal is to provide topic and content suggestions 
to the agent in a non-intrusive manner. This means that the agent can ignore any suggestion that is irrelevant to him/her during the composition of the message.

There are two types of suggestions that we are targeting:
\begin{enumerate}
\item Topic prediction of the entire email response that needs to be composed.
  
  This can be useful for automatically suggesting an appropriate canned response to the agent. If such a document is available, then this can help the agent in planning the response.
  
\item Topic prediction of the next sentence in the reply.
  
  This can be useful for interactively presenting the topics for the next sentence (and the corresponding representative sentence or phrases), which the agent can choose or ignore while composing the reply.
    
\end{enumerate}

We show that, with the methods described in this paper, topic prediction of the entire email response as well as of the next sentence, can be made with reasonably high accuracy, making these predictions potentially useful in a scenario of interactive composition. We note however that these methods could not lead to a fully automatic composition, because, as all techniques that rely only on such textual training data, they do not have access to knowledge bases or similar external resources that the agent needs to consult in order to provide detailed and specific answers.

We evaluated our methods on a set of email exchanges in the Telecom domain in a customer care scenario. Three kinds of experiments were conducted:
\begin{enumerate}
  \item Investigating the influence of the customer's email on the word-level perplexity of the agent's response, to validate and quantify our basic assumption that the context given by the customer's query strongly conditions the agent's response. The details are presented in Section \ref{sec:customer-influence}.
   
   \item Predicting the topics in the agent's response given the customer's query. See Section \ref{sec:overall-email} for details.
   
   \item Predicting the topic(s) in the agent next sentence given the context of his/her previous sentences (in addition to the customer's query). See Section \ref{sec:sentence-level} for details.
\end{enumerate}

%% file: related.tex
\section{Related Work}
\label{sec:related-work}
Most works addressing ``email text analytics'' in the past few years have been on classification and summarization.
Email classification has proven to be useful in many standard applications such as spam detection and filtering high-priority messages. 
Research themes such as summarization and question answering came into focus because of the need of better interpreting the overwhelming amount of emails/messages generated with the advent of email groups and discussion forums.
One of the earlier contributions to email summarization was the work by \cite{Muresan:01} whereas \cite{Rambow:04} extended it to email threads; Scheffer et al. \shortcite{Scheffer:04}  on the other hand proposed semi-supervised classification techniques for question answering in the context of such threads.

Some of the recent classification and summarization techniques have been based on``speech acts'' or ``dialog acts'' such as \emph{proposing a meeting}, \emph{requesting information} \cite{Searle:76,Bunt2011}. Several email studies including summarizing of email threads \cite{Oya:14} or classification of emails \cite{Cohen:04} involve dialog-act based analysis. 
There has been very little work so far on customer-agent related email threads. Some of these works include identification of emotional emails related to customer dissatisfaction/frustration \cite{Gupta:13}, as well as learning possible patterns/phrases for textual re-use in email responses \cite{lamontagne:04}. \cite{Chen2014} is a recent work that attempts to generate emails based on a two-stage process where a structural template is first produced and then a topic-specific language model is used for producing textual realizations of the different slots in the template (see also \cite{Oh2002} for an earlier work using a similar language model based approach).

There have been a number of works that have addressed the problem of discovering the latent structure of topics in the related area of spoken and chat conversations. Recently \cite{Zhai:14} have addressed this problem using HMM-based methods handling dialog state transitions and topic content simultaneously; this work differs from ours in several respects. First, the nature of the data is not the same, short alternating conversational utterances in their case, large single email responses in ours. Second, the focus of \cite{Zhai:14} is on the discovery of latent topics (and conversational states) based on existing dialog texts (speech transcripts or chats), using HMM-based techniques different from our LDA approach. Finally, in our case, in addition to discovering the latent structure of existing emails, we also actually predict which topics will likely be employed in the forthcoming agent's response, which is not attempted in their paper.


%% file: dataset.tex
\section{Dataset}
\label{sec:dataset-description}

The dataset that we used for our study is a collection of emails from the technical support team of a major telecom company in the UK. The dataset contains 54.7k email threads collected from 
the UK region during Jan 2013 $-$ May 2014. Usually, an email thread 
is started by a customer reporting a problem or seeking information, followed 
by an agent's response 
suggesting fixes or asking for more details to fix the reported problem. These threads
continue until the problem is solved or the customer is satisfied with the
agent's response. An example email conversation between a customer and an agent is
given in the left column of Table \ref{tab:example-email}. Usually, customer emails are in free form
while agent replies have a moderately systematic structure. On average, there are 8 emails in a
thread.






\begin{table}[!htbp]
\tabcolsep 2pt
\begin{tabular}{|c|ccc|ccc|}\hline
Type & \multicolumn{3}{c|}{Train} & \multicolumn{3}{c|}{Test}\\
\cline{2-4}
\cline{5-7}
& \smaller{$D$} & \smaller{avg $T$} &\smaller{avg $S$} & \smaller{$D$} & \smaller{avg $T$} & \smaller{avg $S$} \\ \hline
Customer & 38650& 68 &2.6 & 9660& 72 & 2.5\\
Agent & 38650& 221 &8.1& 9660 & 220 &8.0\\
\hline
\end{tabular}
\caption{Statistics of the dataset. D =number of emails, $avg\ T$ = average number of tokens in each email, $avg\ S$ = average number of sentences in email.}
\label{table:sentence-accuracy}
\end{table}

For our study, we just considered the first two emails in a thread, namely the original 
customer's query and the corresponding agent's reply. We have limited 
our experiments to emails which have at least 10 words for customer emails and 20 
words for agent replies. This resulted in {\bf 48.3k email threads} out of which
we used 80\% for training and 20\% for testing. The statistics concerning the number of documents, as well as their average length in tokens (words) and sentences are given in Table \ref{table:sentence-accuracy}.

%% file: silver.tex
\section{Extracting Topics and Building a ``Silver Standard'' based on LDA}
\label{sec:topicmodels}

As we have explained, we focus on two main tasks:
\begin{itemize}
 \item Task \textbf{T1}: predict the likely overall topics of the whole agent's reply  based on the knowledge of the customer's email ;
 \item Task \textbf{T2}:  when an agent is writing an email, predict the likely topics of the next sentence, based on the initial query and the additional knowledge of the previous sentences.
 \end{itemize} 

However, our training data are not annotated at the level of topics. In order to \emph{synthesize} such an annotation, we use a popular unsupervised technique -- Latent Dirichlet Allocation \cite{Blei:03} -- for modeling the topic space of various views of the collection. There are potentially three ways for extracting topics in our set of conversations. In the first setting, we keep customer and agent emails separated, identifying distinct topic models for each collection: a document is a customer email in the first collection, and an agent email in the second collection. We denote by $\mathcal{M}^N_C$ and $\mathcal{M}^M_A$ the topic models trained on customer emails (using $N$ topics) and agent emails (using $M$ topics) respectively. In the second setting, we concatenate customer and agent emails and identify a unique, common set of topics: so, here, we are considering a collection of documents, where each document is the concatenation of the customer's email and of the agent's reply. The resulting model will be noted as $\mathcal{M}^M_{CA}$, where $M$ is the number of topics in the model. In the third setting, instead of considering the whole email as a document, we take each sentence of the email as a separate document when building the topic models. The model is called $\mathcal{M}^M_{S}$. In our case, we built this model from the agent reply messages only, in order to specialize the sentence-level topic model on the agent ```style'' and vocabulary.

As outcome of the topic extraction process on the training sets, we have both the topic distribution over the training documents and the word distribution for each topic. Once trained, from these word distributions and the model priors, we can infer a topic distribution for each word of a given test document (this document being an entire email or a single sentence), and by aggregating these individual distributions, a global  topic distribution can be derived for the whole test document. We consider these assignments of topic distributions as providing a {\bf silver-standard} annotation of the documents (a proxy to a supervised ``gold standard''). These ascribed annotations will subsequently be used for training our prediction models (using the silver-standard annotations of the training set) and evaluating them (using the silver-standard annotations of the test set). Additionally these topic assignments will not only be used as labels to be predicted, but also as additional features to represent and summarize at the semantic level the content of the customer's query (for task \textbf{T1}) and the content of the previous agent sentences for task \textbf{T2} (see details in Section \ref{sec:prediction}).

 We denote a pair of emails of the form customer-query / agent-reply by ($C_i,A_i$) and the application of a topic model $\mathcal{M}^n_t$ to $C_i$ (resp. $A_i$) by $\tau^n_t(C_i)$ (resp. $\tau^n_t(A_i)$), with $n$ the  number of topics and $t$ the type of model ($t \in {C,A,CA,S}$), as described here above. The quantity  $\tau^n_t$ is a probability distribution over $n$ topics and we define the dominant topic $D^n_t$ of a test sample as the topic with the highest probability in this distribution.

In Table \ref{table:topics-sentences}, we present a sample of topics learned using 
the \agentsentences model ($\mathcal{M}^M_{S}$). In the table, we describe each topic with its top
ranked words and phrases.

\begin{table*}[!htbp]
\tabcolsep 2pt
\center
\smaller
\begin{tabular}{cL{8cm}L{8cm}}
\hline
\bf{Topic Label} & \bf{Top words} & \bf{Top phrases}\\ \hline
Contact & support, technical, agent, team , write, contact, & write team,support agent \\
Feedback & contact, enquiry, leave, close, enquiry & answer query, follow link, close enquiry, leave feedback \\
Reset & reset, factory, datum, setting, tap, back, erase, storage & master reset, perform factory reset \\
Repair & repair, device, book, send, centre, email, back, warranty& repair center, book device repair \\
USB & usb, connect, cable, pc, charger, device& disk drive, default connection type, sync manager \\
Cache/App& clear, application, cache, app, datum, setting, delete & clear cache, manage application, cache partition \\
OS/Installation & update, software, system, setting, message, operating & system software update, installation error \\
SD Card & card, sd, account, save, tap, sim, import, people, application& sd card, google account, transfer contact, export sim card \\
Liability & return, charge, liable, un-repaired, device, quote, dispose & hold liable, free charge, choose pay, return handset \\
User Account & tap, account, enter, password, email, setting, step, set, require & username password, email account, secure credentials \\
Damage & repair, charge, brand, return, economic, unrepaired & brand charge, liquid-damaged accessory, return immediately \\ 
SIM/SD Card & card, data, sim, sd, phone, store, device, service, online & test sim card, data remove, insert sim card \\
Settings & tap, scroll, screen, setting, home, icon, notification & home screen, screen tap, notification bar \\
\hline
\end{tabular}
\caption{Representative words and phrases for different topics}
\label{table:topics-sentences}
\end{table*}

\begin{table*}[!htb]
\tabcolsep 2pt
\scalebox{0.70}{
\begin{tabular}{p{16cm}|p{3cm}|p{3cm}|p{3cm}}
\hline
Email Text& Topic1 & Topic2 & Topic3 \\
\hline
\textbf{Customer Query:} My \em{mobile} x fell out of my pocket and the screen 
cracked completely, I was wondering whether I am eligible 
for repair as it is still under the 24 month warranty?&
screen crack, hardware operation, smash screen, display, htc, lcd &
month ago, phone month, buy htc, year ago, phone warranty, contract, contact &
phone work, month ago, time, problem, htc, issue, week, day, back \\
\hline
\textbf{Agent Response:}\normalsize{Thank you for contacting HTC regarding your HTC One X. My name is John 
and I am a Technical Support Agent for the HTC Email Team. I'm sorry to hear that you 
are experiencing difficulties with your device. I understand that the screen is broken. 
Unfortunately this is not covered by warranty, so if you wish to have it repaired, 
you will have to pay a quote. The quotation will be made by the repair centre, and 
it is based on an examination of the handset that is done when it arrives in the repair 
centre. This is why we are unable to provide you with the amount it would cost to have 
the display replaced. I hope that I have given you enough information to solve your query. 
If this is not the case, please do not hesitate to contact us again. If this answer has 
solved your query, and you have no further questions, you can close this ticket by 
clicking on the link shown below. On closing the ticket, you will receive an invitation 
to participate in our Customer Satisfaction Survey. This will only take 1 minute of your time. I wish you a pleasant day.} &
contact htc, write team, support agent, htc regard, technical support, contact htc regard &
repair centre, vary depend exchange rate, physical damage, minor liquid, cover warranty &
leave feedback, close ticket, contact quickly, receive feedback\\
\hline
\end{tabular}}
\caption{Three dominant topics for Customer and Agent, inferred using email text.}
\label{tab:example-email}
\end{table*}

In Table \ref{tab:example-email},  we
present an example of a query/reply email pair, as well as the corresponding Top-3 highest probability topics and their most representative words/phrases, both for the customer and the agent parts ($\mathcal{M}^M_{CA}$ model, with $M = 50$).

\begin{figure}[!htb]
	\centering
	\includegraphics[width=0.48\textwidth, height=50mm]{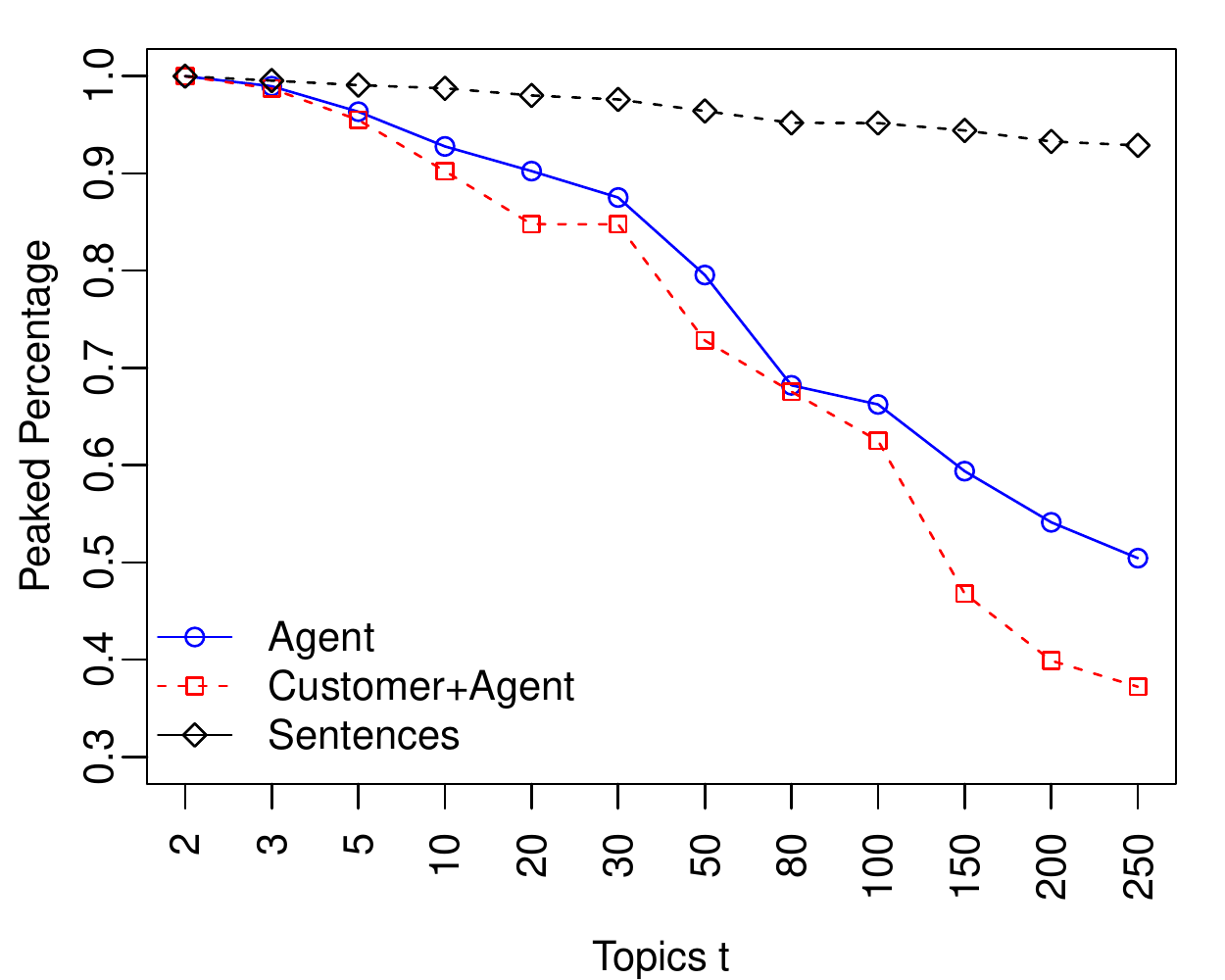}
        \caption{Percentage of sentences that have a ``peaked'' topic distribution).}
        \label{fig:utterance-dominant}	
\end{figure}
Finally, it should be noted that the \agentsentences model ($\mathcal{M}^M_{S}$) gives rise to more peaked topic distributions compared to the other models. We consider that a distribution is peaked when the probability of its dominant topic is more than 0.5 (that is, more than the aggregated probability of all competing topics) and at least twice larger than that of the second topic. Over the test set, more than 90\% of sentences exhibit a peaked distribution when considering the $\tau^M_{S}(A_i)$ topic distribution  (see Figure \ref{fig:utterance-dominant}).  This will motivate the use of the dominant topic instead of the whole distribution when solving the task \textbf{T2}, as explained in Section \ref{sec:sentence-level}.

%% file: perplexity.tex
\section{Influence of the Customer Query on the word-level perplexity}
\label{sec:customer-influence}

We examine the influence of knowing the context of the customer's query on the content of
the agent's email. In order to do that, we consider the set of test agent
emails and compare the perplexity of the language model based on $\mathcal{M}^M_{A}$ versus the
one identified with $\mathcal{M}^M_{CA}$. Recall that the model $\mathcal{M}^M_{A}$ infers the
probability distribution ($\tau_A^M(A_i)$) over topics by exploiting only the agent emails from the training set, while the model $\mathcal{M}^M_{CA}$ infers a
probability distribution ($\tau_{CA}^M(A_i)$) over topics by exploiting both the customer queries and the agent replies in the training set.

The perplexity scores are computed using the following formulas:
\begin{equation}\label{eqn:perplexity}
\small{
\begin{split}
perplexity(\mathbf{A}|\mathcal{M}^M_{A}) = \exp (\frac{- \sum_{i=1}^{d} \log L(A_i)}{\sum_{i=1}^{d} N_{A_i}})
\\
perplexity(\mathbf{A}|\mathbf{C},\mathcal{M}^M_{CA}) = \exp (\frac{- \sum_{i=1}^{d} \log
\frac{L(C_i+A_i)}{L(C_i)}}{\sum_{i=1}^{d} N_{A_i}}) \\
\end{split}
}
\end{equation}

In these equations, the test set is {$((C_1,A_1), (C_2,A_2), \ldots, (C_d,A_d))$}, with $\mathbf{A}=(A_1,\ldots, A_d)$ and $\mathbf{C}=(C_1,\ldots, C_d)$, $d$ is the number of agent emails in the test set and $N_{A_i}$ is the total number of words in $A_i$. The term $L(A_i)$ 
(resp. $L(C_i)$, $L(C_i+A_i)$) 
is the likelihood of the sequence of words in $A_i$
(resp. $C_i$, $C_i+A_i$), as given by the LDA model $\mathcal{M}^M_{A}$ 
(resp. $\mathcal{M}^M_{CA}$, $\mathcal{M}^M_{CA}$).

\begin{figure}[h]
        \centering
        \includegraphics[scale=0.58]{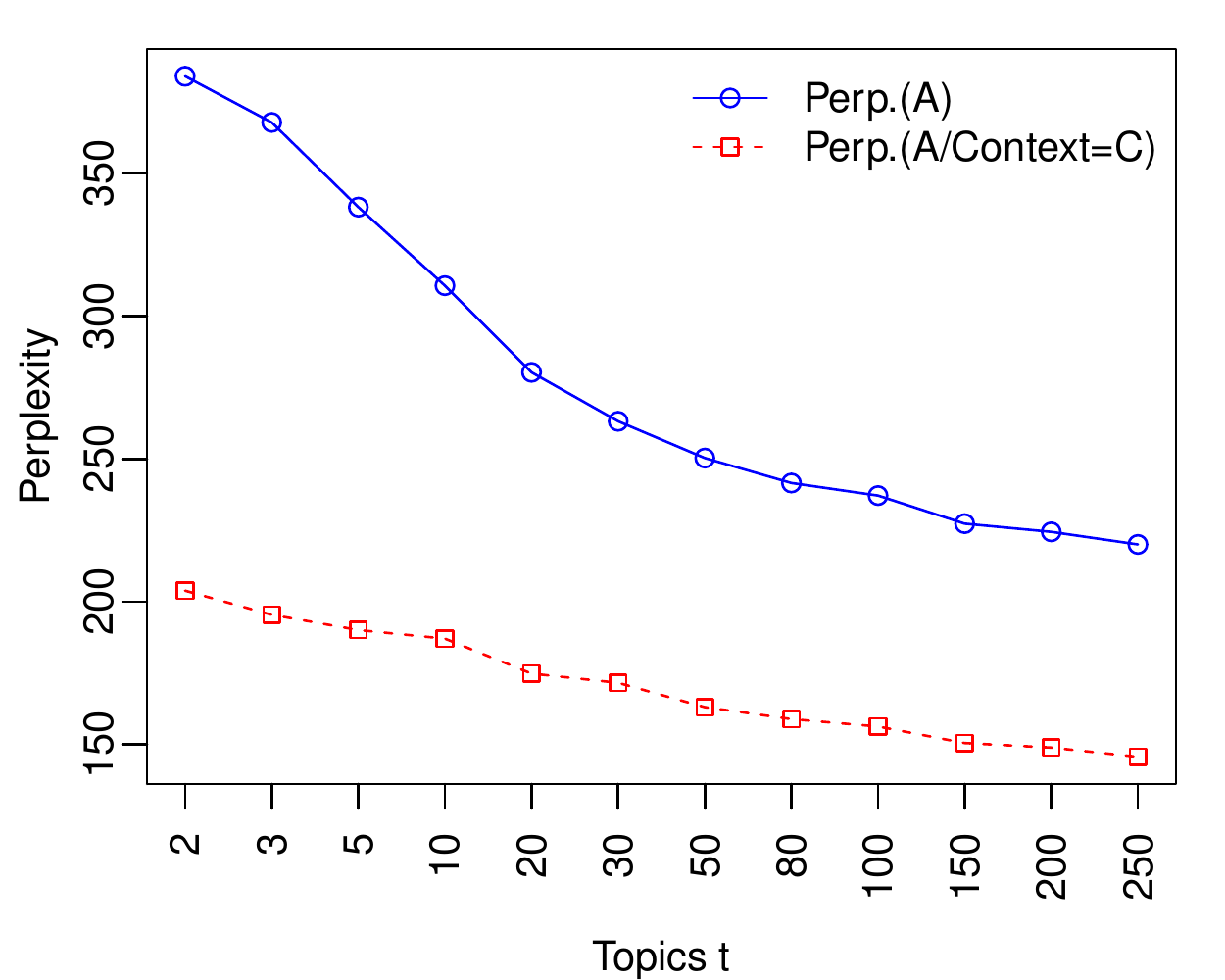}
        \caption{Perplexity}
        \label{fig:perplexity}
\end{figure}

In Figure \ref{fig:perplexity}, we present the perplexity scores of the two models. We see that the model which uses the customer's email as context has lower perplexity scores, as could be expected intuitively. This indicates that a generative LDA model has the potential to use the context $C$ to directly improve the prediction of the \emph{words} in $A$. However, in this paper, instead of directly trying to predict words (which is strongly connected with the design of the user interface, for instance in the form of semi-automatic word completion), we will focus on the different, but related, problem of predicting the most relevant \emph{topics} in a given context. As topics could be rather easily associated with canned responses (sentences, paragraphs or whole emails), predicting the most relevant topics amounts to recommending the most adequate responses.  


%% file: prediction.tex
\section{Predicting Relevant Topics of the Agent Response}
\label{sec:prediction}
\subsection{Topic prediction for the overall agent's email}
\label{sec:overall-email}

In this section, we focus on Task \textbf{T1}, namely predicting the topic distribution of the
agent response using only the contextual information : the customer query $C_i$ and its topic distribution $\tau_{CA}^M(C_i)$. The choice of the $\mathcal{M}^M_{CA}$ model rather than $\mathcal{M}^N_{C}$ is motivated by the considerations described in the previous section (Section \ref{sec:customer-influence}). Note that the $\mathcal{M}^M_{CA}$ model is used both to compute synthetic semantic features ($\tau_{CA}^M(C_i)$) and to provide a silver standard for the topic prediction ($\tau_{CA}^M(A_i)$). The predictor can be written as:

\begin{equation}
	\hat{\tau}^M(A_i) = f(\omega(C_i), \tau_{CA}^M(C_i))
	\label{eqn:logisticregression}
\end{equation}
where $\omega(C_i)$ represents the bag-of-words of the customer query.

Learning the mapping shown in Equation \ref{eqn:logisticregression} could be considered as a structured output learning problem. For solving it, we use an extension of logistic regression that can be trained with soft labels (the silver standard annotations given by $\tau_{CA}^M(A_i)$), adopting the Kullback-Leibler divergence between $\hat{\tau}^M(A_i)$ and $\tau_{CA}^M(A_i)$ as loss function and using a simple Stochastic Descent Gradient algorithm to optimize this loss function. Recall that the silver-standard labels are used both for building the predictor (from the training set) and for assessing the quality of the  predicted topic distribution on the test set.



\subsection{Topic prediction for each sentence of the agent's email}
\label{sec:sentence-level}

To solve our second task (Task \textbf{T2}), namely predicting the topic distribution of the next sentence of an agent's response, we use the words of the customer query $C_i$, its topic distribution $\tau_{CA}^M(C_i)$, the words of the current sentence and the topic distribution of the current sentence $\tau_{S}^M(A_{i,j})$. Note that we are making some kind of Markovian assumption for the agent-side content: we consider that the current sentence and its topic distribution is sufficient to predict the topic of the next sentence, given the customer query context.
Noting $A_{i,j}$ the $j^{th}$ sentence in the agent email $A_i$, we then build the predictor as:
\begin{equation}
\begin{split}
	\hat{\tau}^M_S (A_{i,j+1}) & =\\
	& \hspace*{-3em} f(\omega(C_i),\tau_{CA}^M(C_i),\omega(A_{i,j}),\tau^M_S(A_{i,j}),j))
	\label{eqn:logisticregression2}
\end{split}
\end{equation}
where $j$ is the sentence position (index), and where $\omega(C_i)$ and $\omega(A_i)$ represent the bag-of-words of the customer and agent emails, respectively.

In practice, as we mentioned in Section \ref{sec:topicmodels}, the topic distributions for sentences using the $\mathcal{M}^M_{S}$ models are highly peaked at the ``dominant'' topic. So, it makes sense to use $D^M_S(A_{i,j})$, the dominant topic of the distribution $\tau^M_S(A_{i,j})$ instead of the whole distribution. In the same vein, instead of trying to predict the whole topic distribution of the next sentence, it is reasonable to predict only what will be its dominant topic.  So, a variant of equation \ref{eqn:logisticregression2} is:
\begin{equation}
\begin{split}
	\hat{D}^M_S (A_{i,j+1}) & =\\
	& \hspace*{-3em} f_D(\omega(C_i),\tau_{CA}^M(C_i),\omega(A_{i,j}),D^M_S(A_{i,j}),j))
	\label{eqn:logisticregression3}
\end{split}
\end{equation}
We use the standard multiclass logistic regression for modeling the function shown in
equation \ref{eqn:logisticregression3}. To be more precise, we build $M$ different predictors, one for each possible value of the dominant topic of the current sentence $D^M_S(A_{i,j})$. Moreover, for $j$=0, i.e. for the first sentence, we build a family of simpler ``degenerated'' models, in the following form:
\begin{equation}
	\hat{D}^M_S (A_{i,1}) =
	f_D(\omega(C_i),\tau_{CA}^M(C_i))
	\label{eqn:logisticregression4}
\end{equation}

%% file: evaluation.tex
\section{Evaluation}
\label{sec:evaluation}

In this section, we present experimental results, showing how our proposed methods perform in predicting topics of the agent's response.  
For learning the LDA topic models (as described in section \ref{sec:topicmodels}), we have
used MALLET \cite{Mccallum:02} toolkit, with the standard (default) setting.

We evaluate our methods using three metrics:

\begin{enumerate} 
  \item Bhattacharya coefficient \cite{Bhattacharya:43}

   Here, we evaluate how close the predicted topic distribution  is to the silver-standard topic distribution. For Task \textbf{T1}, we compare 
   $\hat{\tau}^M(A_i)$ with $\tau_{CA}^M(A_i)$ for each agent email $A_i$ of the test set.  For Task \textbf{T2}, we compare $\hat{\tau}^M_S (A_{i,j+1})$ with  $\tau^M_S (A_{i,j+1})$ 
for each sentence $(j+1)$ of the agent emails $A_i$ of the test set. We have also computed more commonly used measures such as KL divergence and they strongly 
correlate our findings with our Bhattacharya coefficient scores.

  \item Text ranking measure (for Task  \textbf{T1})
  
  Instead of directly comparing the probability distributions, we also try
  to measure how useful is the predicted probability distribution in discriminating
  the correct agent's response in comparison to a set of $k$-1 randomly introduced responses 
  from the training set. 
  The $k$ possible answers are ranked according to the Bhattacharya coefficient between their 
  silver-standard topic distribution and the one predicted from the customer's query email following equation \ref{eqn:logisticregression}.
  We consider here the average Recall@1 measure, i.e. the average number of times where the correct response is ranked first.
  
  \item Dominant topic prediction accuracy (for Task  \textbf{T2})
  
   Here, we examine whether the dominant topic of the silver-standard annotation for the next sentence belongs to the top-K predicted topics. A high
   accuracy is necessary for ensuring effective topical suggestions for
   interactive response composition, which is the primary motivation of our
   work.
   
\end{enumerate}

 \subsection{Topic prediction of agent email}

\begin{figure}[!htbp]
        \centering
        \includegraphics[width=0.48\textwidth, bb= 65 245 528 588]{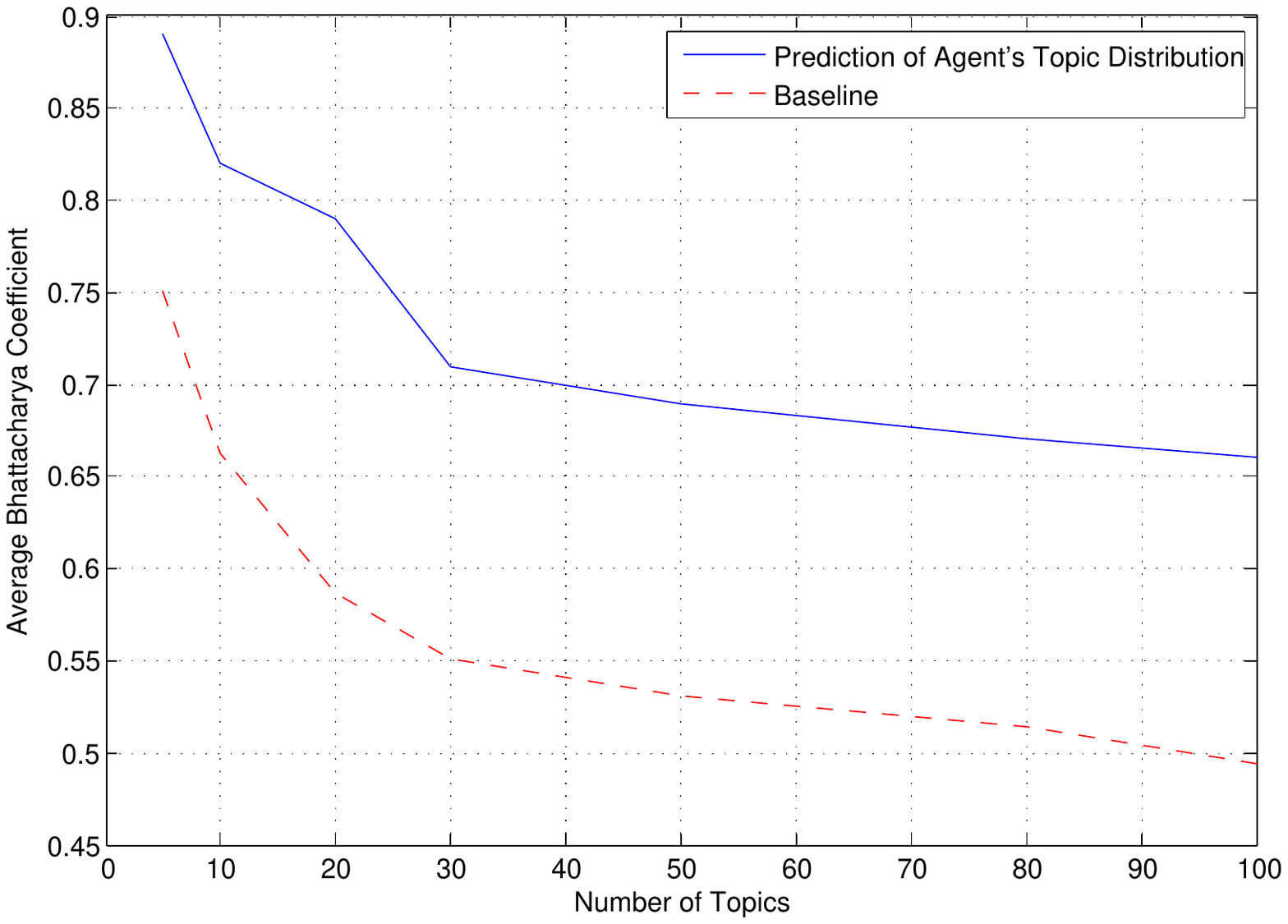}
      \caption{Average Bhattacharya coefficient over all test emails. The solid curve shows the B.~coefficient between the predicted distribution for the agent's email and the silver standard for that email, the dashed curve is a baseline that predicts that the agent's email has the same distribution as the customer's email.}
        \label{fig:avg-bhattacharya}
\end{figure}

In Figure \ref{fig:avg-bhattacharya}, we present the average Bhattacharya
coefficient over all test emails for different numbers of topics ($M$) using the $\mathcal{M}^M_{CA}$ model.\footnote{For our purposes, the Bhattacharya coefficient is preferable to other measures of distribution ``distance'' such as KL-divergence, because it is upper-bounded by $1$, and allows easier comparison across distributions of varying dimensions.}
This figure illustrates the trade-off between the difficulty of the task and the usefulness of the model: a higher number of topics corresponds to a more fine-grained analysis of the content, with potentially better predictive usefulness, but at the same time it is harder to reach a given level of performance (as measured by the Bhattacharya coefficient) than with a small number of topics. For comparison, we also show a baseline where the prediction of the agent's email topic distribution is simply a copy of the customer's email topic distribution, with a much lower performance.

 

\begin{figure}[!htb]
        \centering
        \includegraphics[width=0.48\textwidth, height=50mm]{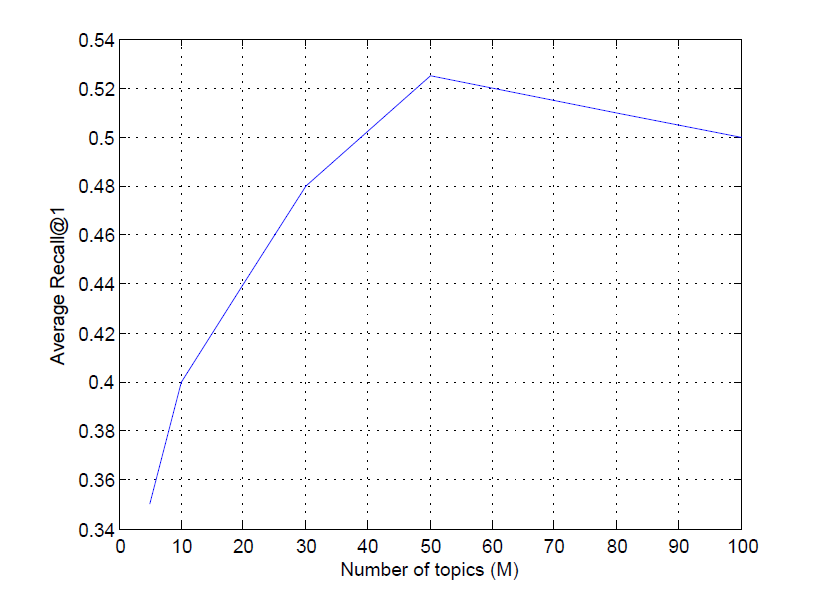} 
        \caption{Text identification accuracy over all test emails.}
        \label{fig:accuracy-bhattacharya}
\end{figure}

Figure \ref{fig:accuracy-bhattacharya} gives the evolution of the average Recall@1 measure for $k=5$ when the number of topics is changed. In this case, the baseline (a simple random guess)  would have given an average Recall@1 equal to 20\%. The  best performance (Recall@1 = 52.5\%) is reached when $M$=50.

\begin{figure}[!htb]
        \centering
        \includegraphics[scale=0.5]{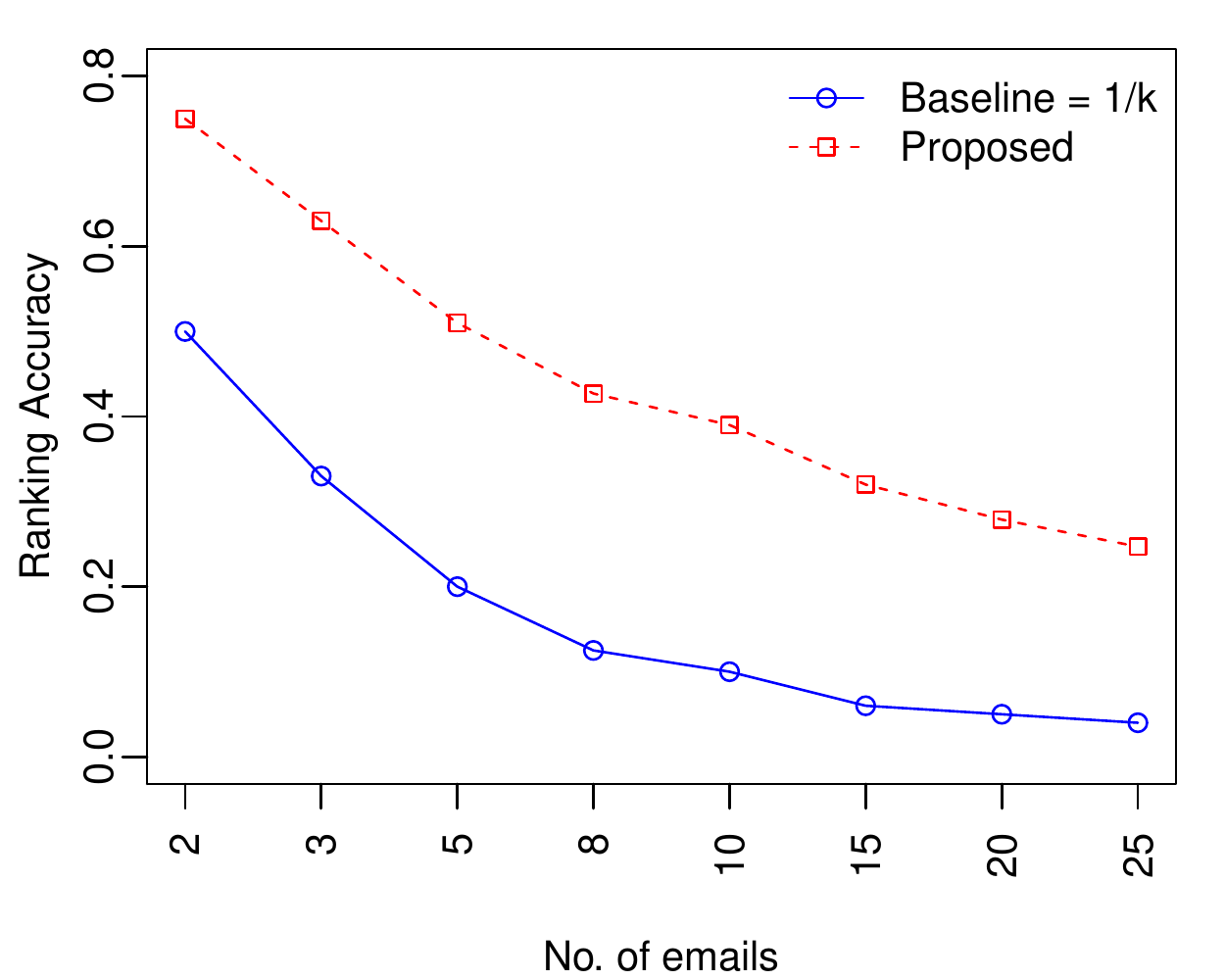}
        \caption{Text identification accuracy with varying $k$.}
        \label{fig:accuracy-bhattacharya-varyingk}
\end{figure}

We have also assessed the average Recall@1 score with varying $k$ and $M$ fixed to 50: see Figure
 \ref{fig:accuracy-bhattacharya-varyingk}. We see that the text ranking based on the topic distribution prediction is always much higher than the baseline scores.

\subsection{Topic Prediction of Next Sentence}
\label{sec:experiments-results}

We compare the dominant topic prediction accuracy of the proposed approach with
other baseline approaches. The baseline approaches that we examined are:

\begin{itemize}
\item \textbf{Uniform:} we assign uniform distribution of topics for every 
sentence in the test set and compare it with silver standard. In other words, we perform a completely random ranking of the topics.

\item \textbf{Average:} we assign the same topic distribution for every sentence of the test set; this topic distribution is the global average topic distribution and can be directly derived from the hyper-parameter $\alpha$ in LDA models by ``normalizing'' the values of the $\alpha$ vector (note that, in our case, this hyper-parameter is learned from the training data).

\end{itemize}



Table  \ref{tab:feature-scores}  gives the dominant topic prediction accuracy for the ``next topic prediction'' task, with $K$=1 (i.e. the relative number of times where the predicted dominant topic corresponds to the dominant topic given by the silver-standard annotation). These values are averaged over all sentences of the test set, irrespective of their position. For this, we have used the predictors given by equation \ref{eqn:logisticregression3} and fixed the number of topics $M$ to 50. Note that the standard multi-class logistic regression outputs a probability distribution over the topics, so that we can compute the Bhattacharya coefficient between this predicted distribution and the silver-standard distribution as well: this information is also given in Table \ref{tab:feature-scores}. To see the relative impact of each type of input features, the table gives the performance for specific subsets of features: the $\omega(x)$ notation represents the bag-of-word feature vector of text entity $x$, while $\tau(x)$ represents the topic distribution vector of text entity $x$.


\begin{table}[h]
\smaller
\tabcolsep 2pt
\begin{tabular}{|cc|cc|cc|cc|}
\hline
\multirow{3}{*}{\textbf{Features}} && \multicolumn{2}{c|}{\textbf{Uniform}} & \multicolumn{2}{c|}{\textbf{Average}} & \multicolumn{2}{c|}{\textbf{Proposed}} \\
\cline{3-5}
\cline{5-6}
\cline{7-8}
&& \smaller{$DTA$} & \smaller{$BC$} &\smaller{$DTA$} & \smaller{$BC$} & \smaller{$DTA$} & \smaller{$BC$} \\ \hline
$\omega(C_i), \omega(A_{i,j})$ && 0.02 & 0.308 &0.064& 0.334 & 0.416 & 0.556 \\
$+position$ && 0.02 & 0.308 & 0.064 & 0.334 & 0.431 & 0.572\\\hline
$\tau(C_i), \tau(A_{i,j})$ && 0.02 & 0.308 & 0.064 & 0.334 & 0.450 & 0.588\\
$+\omega(C_i), \omega(A_{i,j})$ && 0.02 & 0.308 & 0.064 & 0.334 & 0.451 & 0.598\\
+position && 0.02 & 0.308 & 0.064 & 0.334 & \textbf{0.471} & \textbf{0.614} \\
\hline
\end{tabular}
\caption{{Next sentence topic prediction scores, DTA: top-1 Dominant Topic Accuracy, BC: Bhattacharya Coefficient.}}
\label{tab:feature-scores}
\end{table}

We see that the topic prediction accuracy for the next sentence is {\bf 0.471}, which is much higher than both baselines 
(Uniform and Average). We can also see that the topic
distribution vectors of the current sentence and the customer's query give a higher prediction accuracy ({\bf 0.450}) than
using the bag of words features of the context ({\bf 0.416}). When the position of the next sentence is used as a feature,
it improves the results indicating that certain topics are more likely to occur at particular positions in the email than at others. The 
same trend is also seen when we compute the Bhattacharya coefficient (BC) where the predicted topic distribution 
has a much higher BC than both the uniform distribution as well as the Average distribution.

These results (presented in table \ref{tab:feature-scores}) illustrate that about half of the predicted topics match
with the actual topic (based on silver-standard annotations), which is a
significant accuracy in an interactive composition scenario. In Table \ref{tab:topk-feature-scores}, we show the dominant topic prediction accuracies in the top-$K$ predicted topics, with different values of $K$. It can be seen that, in an interactive composition scenario where the agent is presented with 5 recommended topics, the agent will be able to recognize the relevant topic in more than { \bf 80\%} of the cases. When the agent is presented with 2 recommended topics, the agent can choose the right topic in {\bf 62.5\%} of the cases. These results are obtained through a combination of all the features,
with the topic distribution vectors of the context (current sentence and customer's query) playing an important role.

\begin{table}[h]
\smaller
\tabcolsep 2pt
\begin{tabular}{|cc|c|c|c|c|}
\hline
\multirow{2}{*}{\textbf{Features}} && \multicolumn{4}{c|}{\textbf{\smaller{Dominant Topic in top\_K predictions}}}  \\
\cline{3-6}
&& \smaller{$K=1$} & \smaller{$K=2$} &\smaller{$K=5$} & \smaller{$K=10$} \\ \hline
$\omega(C_i), \omega(A_{i,j})$ && 0.416 & 0.532 & 0.722 & 0.851\\
$+position$ && 0.431 & 0.563 & 0.735 & 0.858\\\hline
$\tau(C_i), \tau(A_{i,j})$ && 0.450 & 0.623 & 0.795 & 0.893\\
$+\omega(C_i), \omega(A_{i,j})$ && 0.451 & 0.610 & 0.794 & 0.901\\
$+position$ && \bf{0.471} & \bf{0.625} & {\bf 0.802} & {\bf 0.901}\\
\hline
\end{tabular}
\caption{{Next sentence dominant topic accuracy scores w.r.t top\_K predictions}.}
\label{tab:topk-feature-scores}
\end{table}

%% file: conclusion.tex
\section{Conclusion and Future Work}
\label{sec:future-work}

We have presented new unsupervised models for discovering discourse structures of email replies in customer-agent oriented email systems, and have evaluated their predictive ability on a real-world Contact-Center email dataset, at the global and local levels. Our experiments  indicate the potential of these techniques for an interactive scenario where the agent is guided in the selection of whole emails or individual sentences based on predicted topics. 

Still, numerous interesting avenues could be investigated further. One natural extension of our work would be to consider \emph{Multi-dimensional} LDA models in the sense of \cite{Paul2013}, which are able to detect topics along different semantic aspects, which would be very useful for disentangling several dimensions that we currently do not distinguish, such as: which issue is being talked about, what is the device concerned, at what stage of a conversation we are, and so on.

Another extension would be to examine prediction models that go beyond the Markovian assumption, by exploiting topic dependencies at a longer distance than one sentence. 

The intended application of this work requires developing a user interface, which has implications on the models (level of granularity, number of topics, ...), as well as going beyond the identification of topics towards the interactive generation of actual texts. Ultimately, we want to be able to automatically recognize what is the most relevant response in a given context, not only because it was often given by agents in similar contexts, but because this response will lead to the fastest and more efficient way of solving the customer's problem: we should then couple and solve jointly the topic modelling/prediction task with a sequential optimization problem.